\documentclass[sigconf]{acmart}

\usepackage{alltt}
\usepackage{enumitem}
\usepackage{listings}
\usepackage{xspace}

\newcommand*{\lale}{\textsc{Lale}\xspace}

\definecolor{keyword}{HTML}{37AC4A} 
\definecolor{operator}{HTML}{A51DFF}
\definecolor{string}{HTML}{C03333}
\definecolor{background}{HTML}{F7F7F7}

\definecolor{metamodel}{HTML}{6CA6CD} %skyblue3
\definecolor{planned}{HTML}{7EC0EE} %skyblue2
\definecolor{trainable}{HTML}{B0E2FF} %lightskyblue1

\lstdefinelanguage{schema}{
  basicstyle=\fontsize{8}{9}\sffamily,
  morekeywords={allOf,anyOf,default,description,distribution,enum,maximum,maximumForOptimizer,minimum,not,properties,type,typeForOptimizer},
  morestring=[b]",
}
\newcommand{\schema}[1]{\lstinline[language=schema]{#1}}

\lstdefinelanguage{python}{
  columns=fixed,
  basewidth=0.45em,
  morestring=[b]',
  morestring=[b]""",
  morecomment=[l]\#,
  morekeywords={and,as,assert,break,class,continue,def,del,elif,else,except,False,finally,for,from,global,if,import,in,is,lambda,None,nonlocal,not,or,pass,raise,return,True,try,while,with,yield},
  keywordstyle=\color{keyword}\bf\ttfamily,
  otherkeywords={|,>>,\&,=},
  morekeywords=[2]{|,>>,\&,=},
  keywordstyle=[2]\color{operator}\bf\ttfamily,
}
\newcommand{\python}[1]{\lstinline[language=python]{#1}}

\setcopyright{acmcopyright}
\copyrightyear{\url{https://sites.google.com/view/automl2020-workshop/}}
\acmYear{} %\acmYear{2018}
\acmDOI{} % \acmDOI{10.1145/nnnnnnn.nnnnnnn}

\acmConference[AutoML Workshop @ KDD'20]{AutoML Workshop @ KDD'20}{August 22--27, 2020}{Page~\thepage}
\acmBooktitle{AutoML Workshop @ KDD'20}
\acmPrice{15.00}
\acmISBN{} %\acmISBN{978-1-4503-XXXX-X/18/06}

\startPage{1}
\setcopyright{none}

\begin{document}

\settopmatter{printacmref=false} % Removes citation information below abstract

\title{\lale: Consistent Automated Machine Learning}

\author{Guillaume Baudart, Martin Hirzel, Kiran Kate, Parikshit Ram, and Avraham Shinnar}
\affiliation{
  \institution{IBM Research, USA}}

\begin{abstract}
  Automated machine learning makes it easier for data scientists to
develop pipelines by searching over possible choices for
hyperparameters, algorithms, and even pipeline topologies.
Unfortunately, the syntax for automated machine learning tools is
inconsistent with manual machine learning, with each other, and with
error checks.  Furthermore, few tools support advanced features such
as topology search or higher-order operators. This paper
introduces \lale, a library of high-level Python interfaces that
simplifies and unifies automated machine learning in a consistent way.

\end{abstract}


% \ccsdesc[500]{Computing methodologies~Machine learning}

% \keywords{AutoML, programming models, type checking}

\begin{teaserfigure}
  \includegraphics[width=\textwidth]{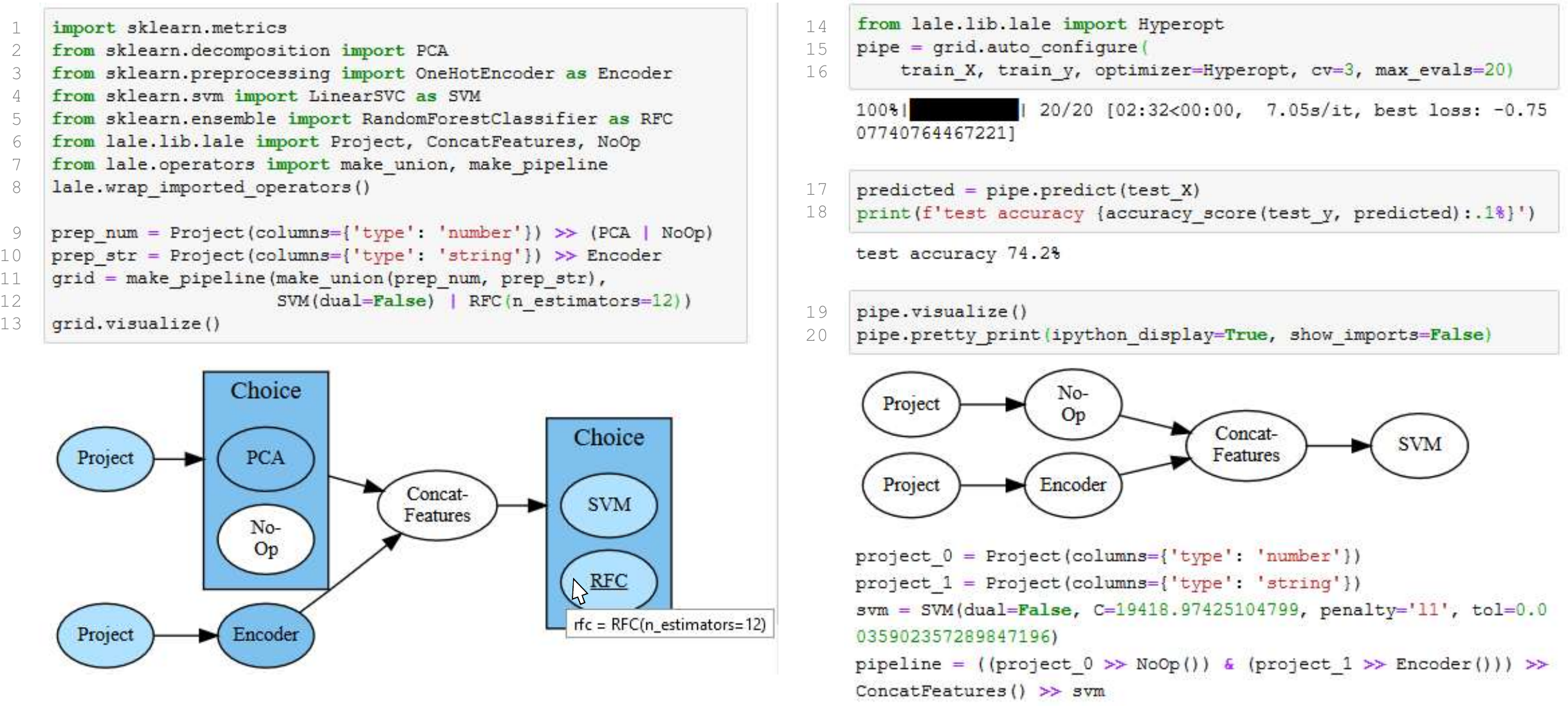}
  \caption{\label{fig:lale_running_example}\lale example for consistent automated machine learning, explained in Section~\ref{sec:language}.}
\end{teaserfigure}

\maketitle

\section{Introduction}\label{sec:intro}

Machine learning (ML) is widely used in various data science problems. There are
many ML {\em operators} for data preprocessing (scaling, missing imputation,
categorical encoding), feature extraction (principal component analysis,
non-negative matrix factorization), and modeling (boosted trees, neural
networks). A machine learning {\em pipeline} consists of one or more operators
that take the input data through a series of transformations to finally
generate predictions. Given the plethora of ML operators (for example, in the
widely used scikit-learn library~\cite{buitinck_et_al_2013}), the task of finding a
good ML pipeline for the data at hand (which involves not only selecting
operators but also appropriately configuring their hyperparameters)
can be tedious and time consuming if done
manually. This has led to a wider adoption of automated machine learning
(AutoML), with the development of novel algorithms (such as
SMAC~\cite{hutter_hoos_leytonbrown_2011}, hyperopt~\cite{bergstra_et_al_2011},
and subsequent recent work~\cite{liu_et_al_2020}), open source libraries
(auto-sklearn~\cite{feurer_et_al_2015},
hyperopt-sklearn~\cite{komer_bergstra_eliasmith_2014,bergstra_et_al_2015},
TPOT~\cite{olson_et_al_2016}), and even commercial tools.

Scikit-learn provides a consistent programming model for manual
ML~\cite{buitinck_et_al_2013}, and various other tools (such as
XGBoost~\cite{chen_guestrin_2016}, LightGBM~\cite{ke_et_al_2017}, and
Tensorflow's \textsf{\small tf.keras.wrappers.scikit\_learn}) maintain consistency with this model
whenever appropriate.  AutoML tools usually feature two pieces -- (i) a way to
define the {\em search space} corresponding to the pipeline topology, the
operator choices in that topology, and their respective possible hyperparameter
configurations, and (ii) an algorithm that explores this search space to
optimize for predictive performance.
Many of these also try to maintain consistency with
the scikit-learn model but only when the user gives up all control and lets the
tool completely automate the ML pipeline configuration, eschewing item~(i)
above. Users often want to retain some control over the
automation, for example, to comply with domain-specific requirements by
specifying operators or hyperparameter ranges to search over. While
fine-grained control over the automation is possible (we will discuss examples),
users need to manually configure the search space. This search space
specification differs for each of the different AutoML tools and
differs from the manual programming model.

We believe that proper \emph{abstractions} are necessary to provide a
consistent programming model across the entire spectrum of controlled
automation. To this end, we introduce \lale, an open-source Python
library\footnotemark, built upon mathematically grounded abstractions of the
elements of the pipeline (for example, topology, operator, hyperparameter
configurations), and designed around scikit-learn~\cite{buitinck_et_al_2013} and
JSON Schema~\cite{pezoa_et_al_2016}. \lale provides a programming interface for
specifying pipelines and search spaces in a consistent manner
while providing fine-grained control across the automation spectrum. The
abstractions also allow us to provide a consistent programming interface for
capabilities {\em for which no such interface currently exists}: (i)~search
spaces with {\em higher-order operators} (operators, such as ensembling,
that have other operators as hyperparameters),
and (ii)~search spaces that include search for the
pipeline topology via {\em context-free grammars}~\cite{chomsky_1956}.
\footnotetext{\url{https://github.com/ibm/lale}}
The contributions of this paper are:

\begin{enumerate}[label=\arabic*.,leftmargin=\parindent]
  \item A pipeline specification syntax that is consistent across the
    automation spectrum and grounded in established technologies.
  \item Automatic search space generation from pipelines and schemas
    for a consistent experience across existing AutoML tools.
  \item Higher-order operators (for ensembles, batching, etc.) with
    automatic AutoML search space generation for nested pipelines.
  \item A grammar syntax for pipeline topology search that is a
    natural extension of the pipeline syntax.
\end{enumerate}

Overall, we hope \lale will make data scientists more productive at
finding pipelines that are consistent with their requirements and
yield high predictive performance.

\section{Problem Statement}\label{sec:problem}

Consistency is a core problem for AutoML and existing libraries fall
short on this front.  This section uses concrete examples from popular
(Auto-)ML libraries to present four shortcomings in existing systems.
We strive to do so in a factual and constructive way.
%We have great admiration for the libraries discussed in this section, and wish to
%make good things even better.

\begin{figure}
\begin{lstlisting}[language=python,frame=single,framerule=0pt,backgroundcolor=\color{background}]
pipe = Pipeline([
    ('transform', PCA(n_components=4)),
    ('classify', LinearSVC(dual=False, C=10))])
pipe.fit(train_X, train_y)
\end{lstlisting}
\caption{\label{fig:example_manual}Example for manual scikit-learn pipeline.}
\end{figure}

\begin{figure}
\begin{lstlisting}[language=python,frame=single,framerule=0pt,backgroundcolor=\color{background}]
pipe = Pipeline([
    ('transform', PCA()),
    ('classify', RandomForestClassifier())])
N = [2, 4, 8]
C = [1, 10, 100, 1000]
param_grid = [
    {'transform__n_components': N,
     'classify': [LinearSVC(dual=False)],
     'classify__C': C},
    {'transform__n_components': N,
     'classify': [RandomForestClassifier(n_estimators=12)]}]
grid = GridSearchCV(pipe, param_grid=param_grid)
grid.fit(train_X, train_y)
\end{lstlisting}
\caption{\label{fig:example_gridsearch}Example for scikit-learn GridSearchCV.}
% https://scikit-learn.org/stable/auto_examples/compose/plot_compare_reduction.html
\end{figure}

\paragraph{$P_1$: Provide a consistent programming model across the automation spectrum.}
There is a spectrum of AutoML ranging from manual machine learning (no
automation) to hyperparameter tuning, operator selection, and pipeline
topology search. Unfortunately, as users progress across this
spectrum, the state-of-the-art libraries require them to learn and use
different syntax and concepts.

Figure~\ref{fig:example_manual} shows an example from the
no-automation end of the spectrum using
scikit-learn~\cite{buitinck_et_al_2013}. The code assembles a two-step
pipeline of a \python{PCA} transformer and a \python{LinearSVC}
classifier and manually sets their hyperparameters, for example,
\python{n_components=4}.

The example in Figure~\ref{fig:example_gridsearch} automates
hyperparameter tuning and operator selection using
\python{GridSearchCV} from scikit-learn. \mbox{Lines 1--3} resemble
Figure~\ref{fig:example_manual}.  \mbox{Lines 4--11} specify a search
space, consisting of a list of two dictionaries. In the first
dictionary, Line~7 specifies the list of values to search over for the
\python{n_components} hyperparameter of the \python{PCA} operator;
Line~8 specifies the classify step of the pipeline to be a
\python{LinearSVC} operator; and Line~9 specifies the list of values
to search over for the \python{C} hyperparameter of the
\python{LinearSVC} operator. The second dictionary is similar, but
specifies the \python{RandomForestClassifier}.

The syntax for a pipeline (Figure~\ref{fig:example_manual} Lines
\mbox{1--3}) differs from that for a search space
(Figure~\ref{fig:example_gridsearch} Lines \mbox{4--11}). The mental
model is that operators and hyperparameters are pre-specified and then
the search space selectively overwrites them with different choices.
To do so, the code uses strings to name steps and hyperparameters,
with a double underscore (\python{'__'}) name mangling convention to
connect the two.  Relying on strings for names can cause
hard-to-detect mistakes~\cite{baudart_et_al_2019}.
In contrast, using a single syntax for both manual
and automated pipelines would make them more consistent and would
obviate the need for mangled strings to link between the two.

\paragraph{$P_2$: Provide a consistent programming model across different AutoML tools.}
Compared to \python{GridSearchCV}, Bayesian optimizers such as
hyperopt-sklearn~\cite{komer_bergstra_eliasmith_2014} and
auto-sklearn~\cite{feurer_et_al_2015} speed up search using
smarter search strategies. Unfortunately, each of these AutoML tools
comes with its own syntax and concepts.

\begin{figure}
\begin{lstlisting}[language=python,frame=single,framerule=0pt,backgroundcolor=\color{background}]
N = scope.int(hp.qloguniform('N', 2, 8, 1))
C = hp.lognormal('C', 1, 1000)
estim = HyperoptEstimator(
    preprocessing=[pca('transform', n_components=N)],
    classifier=hp.choice('classify', [
        liblinear_svc('classify.svc', dual=False, C=C),
        random_forest('classify.rf', n_estimators=12)]),
    algo=tpe.suggest, max_evals=100, trial_timeout=120)
estim.fit(train_X, train_y)
\end{lstlisting}
\caption{\label{fig:example_hyperopt}Example for hyperopt-sklearn.}
\end{figure}

Figure~\ref{fig:example_hyperopt} shows an example
%that automates
%hyperparameter tuning and operator selection
using the
hyperopt-sklearn~\cite{komer_bergstra_eliasmith_2014} wrapper for
hyperopt~\cite{bergstra_et_al_2011}. Line~1 specifies a discrete
search space \python{N} with a logarithmic prior, a range from
\mbox{2..8}, and a quantization to multiples of~1. Line~2 specifies a
continuous search space \python{C} with a logarithmic prior and a
range from \mbox{1..1000}. Line~4 sets the transform step of the
pipeline to \python{pca} with hyperparameter
\python{n_components=N}. \mbox{Lines 5--7} set the classify step
% of the pipeline
to a choice between \python{linear_svc} and
\python{random_forest}.
%with their hyperparameters.

\begin{figure}
\begin{lstlisting}[language=python,frame=single,framerule=0pt,backgroundcolor=\color{background}]
estim = AutoSklearnClassifier(
    include_preprocessors=['pca'],
    include_estimators=['liblinear_svc', 'random_forest'],
    time_left_for_this_task=1800, per_run_time_limit=120)
estim.fit(train_X, train_y)
\end{lstlisting}
\caption{\label{fig:example_autosklearn}Example for auto-sklearn based on SMAC.}
\end{figure}

Figure~\ref{fig:example_autosklearn} shows the same example using
auto-sklearn~\cite{feurer_et_al_2015}. While power users can use
the
\python{ConfigurationSpace} used by
SMAC~\cite{hutter_hoos_leytonbrown_2011} to
adjust search
spaces for individual hyperparameters, we elide this for
brevity. Line~2 sets the preprocessor to \python{'pca'} and Line~3 sets the
classifier to a choice of \python{'linear_svc'} or \python{'random_forest'}.

The syntaxes for the three AutoML tools in Figures
\ref{fig:example_gridsearch}, \ref{fig:example_hyperopt},
and~\ref{fig:example_autosklearn} differ. There are three ways to
refer to the same operator: \python{PCA}, \python{pca(..)}, and
\python{'pca'}. There are three ways to specify an operator choice: a
list of dictionaries, \python{hp.choice}, and a list of strings.  The
mental model varies from overwriting to nested configuration to
string-based configuration. Users must learn new syntax and concepts
for each tool and must rewrite code to switch tools.  Moreover, as we
consider more sophisticated pipelines (beyond the simple two-step one
presented in the example), the search space specifications get even
more convoluted and diverse between these existing specification
schemes.  A unified syntax would make these tools
more consistent, easier to learn, and easier to switch. Furthermore,
this syntax should unify not just AutoML tools but also be consistent
with the manual end of the spectrum ($P_1$). More specifically, given
that scikit-learn sets the de-facto standard for manual ML, the syntax
should be scikit-learn compatible.

\paragraph{$P_3$: Support topology search and higher-order operators in AutoML tools.}
The tools previously discussed search operators and hyperparameters
but do not optimize the topology of the pipeline itself. There are
some tools that do, including TPOT~\cite{olson_et_al_2016},
Recipe~\cite{desa_et_al_2017}, and
AlphaD3M~\cite{drori_et_al_2019}. Unfortunately, their methods for
specifying the search space are inconsistent
with manual machine learning and established tools.  TPOT does not
allow the user to specify the search space for pipeline
topologies (the user can specify the set of operators and can fix the pipeline topology, disabling the
topology search). Recipe and AlphaD3M use context-free grammars to
specify the search space for the topology, but in a manner 
inconsistent with each other or with other levels of automation.

\begin{figure}
\includegraphics[width=\columnwidth]{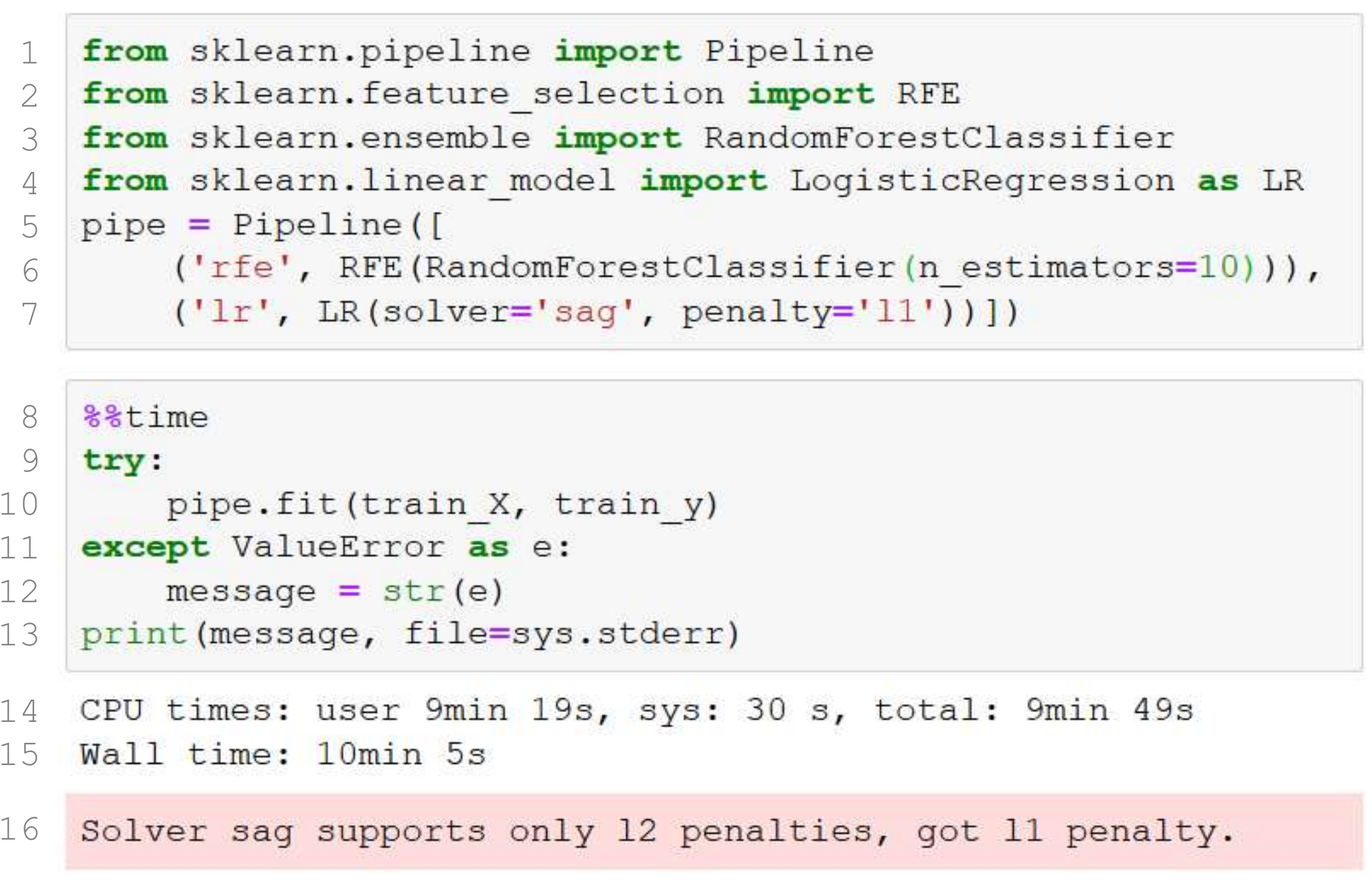}
\caption{\label{fig:sklearn_error_check}Example of scikit-learn error checking.}
\end{figure}

Some transformers (e.g.~\python{RFE} in
Figure~\ref{fig:sklearn_error_check} Line~7) and estimators
(e.g.~\python{AdaBoostClassifier}) are \emph{higher-order}: they take
other operators as arguments. Using the 
AutoML tools discussed so far to search inside their nested operators
is not straightforward. The
aforementioned TPOT, Recipe, and AlphaD3M do not handle higher-order operators in
their search for pipeline topology.

A unified syntax for topology search and higher-order operators
that is a natural extension of the syntax for manual machine learning,
algorithm selection, and hyperparameter tuning would make AutoML more
expressive while keeping it consistent.

\paragraph{$P_4$: Check for invalid configurations early and prune them out of search spaces.}
Even if the search for each hyperparameter uses a valid range in
isolation, their combination can violate side constraints. Worse,
these errors may be detected late, wasting time.

Figure~\ref{fig:sklearn_error_check} shows a misconfigured pipeline:
the hyperparameters for \python{LR} in Line~7 are valid in
isolation but invalid in combination. Unfortunately, this is not
detected when the pipeline is created on Line~7. Instead, when
Line~10 tries to fit the pipeline, it first fits the
first step of the pipeline (see \python{RFE} in Line~6). Only then
does it try to fit the \python{LR} and detect
the mistake. This wastes 10~minutes (Line~15).

In contrast, a declarative specification of the side constraints would
both speed up this manual example and enable AutoML search space
generators to prune cases that violate constraints, thus speeding up
the automated case too. Furthermore, in some situations, invalid
configurations cause more harm than just wasted time, leading
optimizers astray (Section~\ref{sec:results}). It is possible (with
varying levels of difficulty) to incorporate these side constraints
with the search space specification schemes used by the tools
discussed earlier, but they each have inconsistent methods
for doing this.  Moreover, the complexity of these specifications
make them error prone.
%allowing
%some misconfigurations to still go through (while pruning valid
%configurations).
Additionally, while these side constraints
help the optimizer, they do not directly help detect misconfigurations (as
in Figure~\ref{fig:sklearn_error_check}).  Custom
validators would need to be written for each tool.

% \medskip
%The problem statement for this paper is \mbox{$P_1\wedge P_2\wedge
%  P_3\wedge P_4$}.

%%% Local Variables:
%%% mode: latex
%%% TeX-master: "main"
%%% End:

\section{Programming Abstractions}\label{sec:language}

This section shows \lale's abstractions for consistent AutoML,
addressing the problem statements \mbox{$P_1\wedge P_2\wedge P_3\wedge
  P_4$} from Section~\ref{sec:problem}.

\subsection{Abstractions for Declarative AutoML}\label{sec:lang_declarative}

An \emph{individual operator} is a data science algorithm (aka.\ a
primitive or a model), which may be a transformer or an estimator such
as classifier or a regressor. Individual operators are modular
building blocks from libraries such as scikit-learn.
Figure~\ref{fig:lale_running_example} Lines 2--6 contain several
examples, e.g., \python{import LinearSVC as SVM}. Mathematically,
\lale views an individual operator as a function of the form
\[\mathit{indivOp} : \theta_\text{hyperparams}\to\mathcal{D}_\text{fit}\to\mathcal{D}_\text{in}\to\mathcal{D}_\text{out}\]

\noindent
This view uses \emph{currying}: it views an operator as a sequence of
functions each with a single argument and returning the next in the
sequence. An individual operator (e.g., \python{SVM}) is a function
from hyperparameters $\theta_\text{hyperparams}$ (e.g.,
\python{dual=False}) to a function from training data $\mathcal{D}_\text{fit}$
(e.g., \python{train_X, train_y}) to a function from input data
$\mathcal{D}_\text{in}$ (e.g., \python{test_X}) to output data $\mathcal{D}_\text{out}$
(e.g., \python{predicted} in Figure~\ref{fig:lale_running_example}
Line~17). In the beginning, all arguments are latent, and each step in
the sequence captures the next argument as given. The scikit-learn
terminology for the three curried sub-functions is init, fit, and
predict. Viewing operators as mathematical functions avoids
complications arising from in-place mutation. It lets us conceptualize
\emph{bindings as lifecycle}: each bound, or captured, argument
unlocks the functionality of the next state in the lifecycle.

A \emph{pipeline} is a directed acyclic graph (\emph{DAG}) of
operators and a pipeline is itself also an operator. Since a pipeline
contains operators and is an operator, it is highly composable.
Furthermore, viewing both individual operators and pipelines as
special cases of \emph{operators} makes the concepts more
consistent. An example is
\python{make_pipeline(make_union(PCA, RFC), SVM)}, which is equivalent
to
\python{((PCA & RFC) >> ConcatFeatures >> SVM)}.
Here, \python{&} is the \emph{and combinator} and \python{>>} is the
\emph{pipe combinator}. Combinators make edges more explicit and code
more concise. An expression \python{x & y} composes \python{x} and
\python{y} without introducing additional edges. An expression
\python{x >> y} introduces edges from all sinks of subgraph \python{x}
to all sources of~\python{y}. Mathematically, \lale views a pipeline
as a function of the form
\[\mathit{pipeline} : \theta_\text{topology}\to\theta_\text{hyperparams}\to\mathcal{D}_\text{fit}\to\mathcal{D}_\text{in}\to\mathcal{D}_\text{out}\]

\noindent
This uses currying just like individual operators, plus an additional
$\theta_\text{topology}$ at the start to capture the steps and edges.
A pipeline is \emph{trainable} if both $\theta_\text{topology}$ and
$\theta_\text{hyperparams}$ are given, i.e., the hyperparameters of
all steps have been captured.  To fit a trainable pipeline, iterate
over the steps in a topological order induced by the edges. For each
step $s$, let $\mathcal{D}_\text{fit}^s$ be the training data for the step, which is
either the pipeline's training data if $s$ is a source or the
predecessors' output data
$\mathcal{D}_\text{fit}^s=[\mathcal{D}_\text{out}^p]_{p\in\mathit{preds}(s)}$ otherwise.  Then,
recalling that $s$ is a curried function, calculate
$s_\text{trained}=s(\mathcal{D}_\text{fit}^s)$ and
$\mathcal{D}_\text{out}^s=s_\text{trained}(\mathcal{D}_\text{fit}^s)$. The trained pipeline
substitutes trained steps for trainable steps in
$\theta_\text{topology}$. To make predictions with a trained pipeline,
simply interpret \python{>>} as function composition~$\circ$.

An \emph{operator choice} is an exclusive disjunction of operators and
is itself also an operator. Operator choice specifies algorithm
selection, and by being an operator, addresses problem $P_1$ from
Section~\ref{sec:problem}. An example is \python{(SVM | RFC)}, where
\python{|} is the \emph{or combinator}. Mathematically, \lale views an
operator choice as a function of the form
\[\mathit{opChoice} : \theta_\text{steps}\to\theta_\text{hyperparams}\to\mathcal{D}_\text{fit}\to\mathcal{D}_\text{in}\to\mathcal{D}_\text{out}\]

\noindent
This again uses currying. Argument $\theta_\text{steps}$ is the list
of operators to choose from. The $\theta_\text{hyperparams}$ of an
operator choice consists of an indicator for which of its steps is
being chosen, along with the hyperparameters for that chosen
step. Once $\theta_\text{hyperparams}$ is captured, the operator
choice is equivalent to just the chosen step, as shown in the
visualization after Figure~\ref{fig:lale_running_example} Line~20.

\begin{figure}
\begin{lstlisting}[language=schema,frame=single,framerule=0pt,backgroundcolor=\color{background}]
PCA: {
  type: object,
  properties: {
    N: { anyOf: [
      { description: "Amount of variance to explain",
        type: number, minimum: 0.0, maximum: 1.0,
        distribution: uniform},
      { description: "Guess with Minka's MLE", enum: [mle]}]}}}
J48: { allOf: [
  { type: object,
    properties: {
      R: { description: "Use reduced error pruning", type: boolean },
      C: { description: "Pruning confidence threshold",
           type: number, minimum: 0.0, maximum: 1.0,
           maximumForOptimizer: 0.5, distribution: uniform }}},
  { description: "Setting confidence makes no sense for R",
    anyOf: [
      { not: { type: object, properties: {R: {enum: [true]}}}},
      { type: object, properties: {C: {enum: [0.25]}}}]}]}
LR: { allOf: [
  { type: object,
    properties: {
      S: { description: "Optimization problem solver",
           enum: [linear, sag, lbfgs], default: linear},
      P: { description: "Penalization norm",
           enum: [l1, l2], default: l2}}},
  { description: "Solvers sag and lbfgs support only l2.",
    anyOf: [
      { not: { type: object, properties: {S: {enum: [sag, lbfgs]}}}},
      { type: object, properties: {P: {enum: [l2]}}}]}]}
\end{lstlisting}
\caption{\label{fig:example_hpschemas}JSON Schemas for hyperparameters.}
\end{figure}

The \emph{combined schema} of an operator specifies the valid values
along with search guidance for its latent arguments. It addresses
problem $P_4$ from Section~\ref{sec:problem}, supporting automated
search with a pruned search space and early error checking all from
the same single source of truth. Consider the pipeline
\mbox{\python{PCA >> (J48 | LR)}}, where \python{PCA} and \python{LR}
are the principal component analysis and logistic regression from
scikit-learn and J48 is a decision tree with pruning from
Weka~\cite{hall_et_al_2009}.  These operators have many
hyperparameters and constraints and \lale handles all of them. For
didactic purposes, this section discusses only a representative
subset. Figure~\ref{fig:example_hpschemas} shows the JSON
Schema~\cite{pezoa_et_al_2016} specification of that subset.
The open-source \lale library includes JSON schemas for many
operators, some hand-written and others
auto-generated~\cite{baudart_et_al_2020}.
The number of components for PCA is given by $N$, which can be a
continuous value in~$(0..1)$ or the categorical value \textit{mle}.
The prior \schema{distribution} helps AutoML tools search faster.
J48 has a categorical hyperparameter $R$ to enable reduced error
pruning and a continuous confidence threshold $C$ for pruning.
Mathematically, we denote \schema{allOf} as~$\wedge$, \schema{anyOf}
as~$\vee$, and \schema{not} as~$\neg$. Even though the valid values
for $C$ are $(0..1)$, search should only consider values
$C\in(0..0.5)$.  The constraint in \mbox{Lines 16--19} encodes a
conditional \mbox{$(R=\textit{true})\Rightarrow(C=0.25)$} by using the
equivalent \mbox{$\neg(R=\textit{true})\vee(C=0.25)$}.
LR has two categorical hyperparameters $S$ (solver) and $P$ (penalty),
with a constraint that solvers \textit{sag} and \textit{lbfgs} only
support penalty \textit{l2}, which we already encountered in
Figure~\ref{fig:sklearn_error_check}.
Going forward, we will denote a JSON Schema \schema{object} with
\schema{properties} as $\textrm{dict}\texttt{\{\}}$ and a JSON Schema
\schema{enum} as~$[\;]$. Eliding priors, this means
Figure~\ref{fig:example_hpschemas} becomes:
\[\begin{array}{l@{\,}c@{\;}l}
  \textit{PCA}
& : & \textrm{dict}\texttt{\{}N{:\,}(0..1) \vee [\textit{mle}])\texttt{\}}
\\
  \textit{J48}
& : & \textrm{dict}\texttt{\{}
        R{:\,}[\textit{true},\textit{false}],
        C{:\,}(0..0.5)\texttt{\}} \wedge\\
&   & ( \textrm{dict}\texttt{\{}R{:\,}[\textit{true}]\texttt{\}} \Rightarrow
        \textrm{dict}\texttt{\{}C{:\,}[0.25]\texttt{\}})
\\
  \textit{LR}
& : & \textrm{dict}\texttt{\{}
        S{:\,}[\textit{linear}, \textit{sag}, \textit{lbfgs}],
        P{:\,}[\textit{l1}, \textit{l2}]\texttt{\}} \wedge\\
&   &  ( \textrm{dict}\texttt{\{}S{:\,}[\textit{sag}, \textit{lbfgs}]\texttt{\}} \Rightarrow
         \textrm{dict}\texttt{\{}P{:\,}[\textit{l2}]\texttt{\}})
\end{array}\]

To \emph{auto-configure} an operator means to automatically capture
$\theta_\text{hyperparams}$, which involves jointly selecting
algorithms (for operator choices) and tuning hyperparameters (for
individual operators). We saw examples for doing this with
scikit-learn's GridSearchCV~\cite{buitinck_et_al_2013},
hyperopt-sklearn~\cite{komer_bergstra_eliasmith_2014}, and
auto-sklearn~\cite{feurer_et_al_2015} in Figures
\ref{fig:example_gridsearch}, \ref{fig:example_hyperopt},
and~\ref{fig:example_autosklearn}. \lale offers a single unified
syntax shown in Figure~\ref{fig:lale_running_example} \mbox{Lines
  15--16} to address problem $P_2$ from Section~\ref{sec:problem}.
Mathematically, let \textit{op} be either an individual operator or a
pipeline or choice whose $\theta_\text{topology}$ or
$\theta_\text{steps}$ are already captured.  That means \textit{op}
has the form $\theta_\text{hyperparams}\to\mathcal{D}_\text{fit}\to
\mathcal{D}_\text{in}\to\mathcal{D}_\text{out}$. Then
$\textit{auto\_configure}(\textit{op}, \mathcal{D}_\text{fit})$
returns a function of the form
\mbox{$\mathcal{D}_\text{in}\to\mathcal{D}_\text{out}$}.
Section~\ref{sec:compiler} discusses how to implement
\textit{auto\_configure} using schemas.

\subsection{Abstractions for Controlled AutoML}\label{sec:lang_controlled}

AutoML users rarely want to completely surrender all decisions to the
automation. Instead, users typically want to control certain decisions
based on their problem statement and expertise. This section discusses
how \lale supports such \emph{controlled} AutoML.

The \emph{lifecycle state} of an operator is the number of curried
arguments it has already captured that determines the functionality it
supports.  To enable controlled AutoML, \lale pipelines can mix
operators with different states. The visualizations in
Figure~\ref{fig:lale_running_example} indicate lifecycle states via
colors.  A \emph{planned} operator, shown in \colorbox{planned}{dark
  blue}, has captured only $\theta_\text{topology}$ or
$\theta_\text{steps}$, but $\theta_\text{hyperparams}$ is still
latent. Planned operators support \python{auto_configure} but not
\python{fit} or \python{predict}. A \emph{trainable} operator, shown
in \colorbox{trainable}{light blue}, has also captured
$\theta_\text{hyperparams}$, leaving $\mathcal{D}_\text{fit}$ 
latent. Trainable operators support \python{auto_configure} and
\python{fit} but not \python{predict}. A \emph{trained} operator,
shown in white, also captures $\mathcal{D}_\text{fit}$. Trained operators support
\python{auto_configure}, \python{fit}, and \python{predict}. Later
states subsume the functionality of earlier states, enabling user
control over where automation is applied. The state
of a pipeline is the least upper bound of the states of its steps.
%: if
%all steps are trained, the pipeline is trained, otherwise if all steps
%are trainable, the pipeline is trainable, otherwise it is planned.

\emph{Partially captured hyperparameters} of an individual operator
treat some of its hyperparameters as given while keeping the remaining
ones latent. Specifying some hyperparameters by hand lets users control
and hand-prune the search space while still searching other
hyperparameters automatically. For example,
Figure~\ref{fig:lale_running_example} Line~12 shows partially captured
hyperparameters for \python{SVM} and \python{RFC}. The visualization
after Line~13 reflects this in a tooltip for \python{RFC}. And the
pretty-printed code after Line~20 shows that automation respected the
given \python{dual} hyperparameter of \python{SVM} while capturing the
latent \python{C}, \python{penalty}, and \python{tol}. Individual
operators with partially captured hyperparameters are trainable:
\python{fit} uses defaults for latents.

\emph{Freezing} an operator turns future \python{auto_configure} or
\python{fit} operations into an identity.
\python{PCA(n_components=4).freeze_trainable() >> SVM} freezes all
hyperparameters of \python{PCA} (using defaults for its latents), so
\python{auto_configure} on this pipeline tunes only the
hyperparameters of \python{SVM}. Similarly, on a trained operator,
\python{freeze_trained()} freezes its learned coefficients, so any
subsequent \python{fit} call will ignore its
new~$\mathcal{D}_\text{fit}$.  Freezing part of a pipeline speeds up
(Auto-)ML.

\begin{figure}
\begin{lstlisting}[language=python,frame=single,framerule=0pt,backgroundcolor=\color{background}]
from xgboost import XGBRegressor as Forest
import lale.schemas as schemas
lale.wrap_imported_operators()
Grove = Forest.customize_schema(
    n_estimators=schemas.Int(min=2, max=6),
    booster=schemas.Enum(['gbtree']))
\end{lstlisting}
\caption{\label{fig:example_grove}Example for custom schemas.}
\end{figure}

A \emph{custom schema} derives a variant of an individual operator
that differs only in its schema. Custom schemas can modify ranges or
distributions for search. As an extreme case, users can attach custom
schemas to non-\lale operators to enable hyperparameter tuning on
them---the call to \python{wrap_imported_operators} in
Figure~\ref{fig:lale_running_example} Line~8 implicitly does that.
Figure~\ref{fig:example_grove} shows how to explicitly customize a
schema. Line~2 imports helper functions for expressing JSON Schema in
Python. XGBoost implements a forest of boosted trees, so to get a small forest
(a \python{Grove}), Line~5 restricts the number of trees to $[2..6]$.  Line~6
restricts the \python{booster} to a constant, using a singleton
enum. Afterwards, \python{Grove} is an individual operator like any
other. Mathematically, we can view the ability to attach a schema to an
individual operator as extending its curried function on the left:
\[\mathit{indivOp} : \theta_\text{schemas}\to\theta_\text{hyperparams}\to\mathcal{D}_\text{fit}\to\mathcal{D}_\text{in}\to\mathcal{D}_\text{out}\]

\subsection{Abstractions for Expressive AutoML}\label{sec:lang_expressive}

\begin{figure}
\begin{lstlisting}[language=python,frame=single,framerule=0pt,backgroundcolor=\color{background}]
pipeline = make_pipeline(
    MinMaxScaler | StandardScaler | Normalizer
        | RobustScaler | QuantileTransformer,
    AdaBoostClassifier(base_estimator=DecisionTreeClassifier))
\end{lstlisting}
\caption{\label{fig:example_adaboost}Example for higher-order operator.}
\end{figure}

A \emph{higher-order operator} is an operator that takes another
operator as an argument. Scikit-learn includes several higher-order
operators including \python{RFE}, \python{AdaBoostClassifier}, and
\python{BaggingClassifier}. The nested operator is a hyperparameter of
the higher-order operator and the nested operator can have
hyperparameters of its own. Finding the best predictive performance
requires AutoML to search both outside and inside higher-order
operators.  Figure~\ref{fig:example_adaboost} shows an example
higher-order \python{AdaBoostClassifier} with a nested
\python{DecisionTreeClassifier}. On the outside, AutoML should search
the operator choice in \mbox{Lines 2--3} and the
hyperparameters of \texttt{\fontsize{8}{9}\ttfamily AdaBoost\-Classifier} such as
\python{n_estimators}. On the inside, AutoML should
tune the hyperparameters of \python{DecisionTreeClassifier}.  \lale
searches both jointly, helping solve problem $P_3$ from
Section~\ref{sec:problem}. The JSON schema of the
\python{base_estimator} hyperparameter of \python{AdaBoostClassifier}
is:
\begin{lstlisting}[language=schema,frame=single,framerule=0pt,backgroundcolor=\color{background}]
{ description: "Base estimator from which the ensemble is built.",
  anyOf: [
    {typeForOptimizer: "operator"},
    {enum: [null]}],
  default: null}
\end{lstlisting}

A \emph{pipeline grammar} is a context-free grammar that describes a
possibly unbounded set of pipeline topologies. A grammar describes a
search space for the $\theta_\text{topology}$ and
$\theta_\text{steps}$ arguments needed to create planned pipelines and
operator choices. Grammars formalize AutoML tools for topology search
such as TPOT~\cite{olson_et_al_2016}, Recipe~\cite{desa_et_al_2017},
and AlphaD3M~\cite{drori_et_al_2019} that capture
$\theta_\text{topology}$ and $\theta_\text{steps}$ automatically.
\lale provides a grammar syntax that is a natural extension of
concepts described earlier, helping solve problem $P_3$ from
Section~\ref{sec:problem}.

\begin{figure}
\begin{lstlisting}[language=python,frame=single,framerule=0pt,backgroundcolor=\color{background}]
g = Grammar()

g.start  = g.est | (g.clean >> g.est) | (g.tfm >> g.est) \
         | (g.clean >> g.tfm >> g.est)
g.clean  = (g.clean1 >> g.clean) | g.clean1
g.tfm    = (g.tfm1 >> g.tfm) | g.tfm1

g.clean1 = SimpleImputer | MissingIndicator
g.tfm1   = PCA | OrdinalEncoder \
         | OneHotEncoder(handle_unknown='ignore')
g.est    = GaussNB | RidgeClassifier | LinearSVC | SGDClassifier

grid = g.unfold(3)
pipe = grid.auto_configure(
    train_X, train_y, optimizer=Hyperopt, cv=3, max_evals=100)
\end{lstlisting}
\caption{\label{fig:example_alphad3m}Example grammar inspired by AlphaD3M.}
\end{figure}

Figure~\ref{fig:example_alphad3m} shows a \lale grammar inspired by
the AlphaD3M paper~\cite{drori_et_al_2019}. It describes linear
pipelines comprising zero or more data cleaning operators, followed by
zero or more transformers, followed by exactly one estimator.  This is
implemented via recursive non-terminals: \python{g.clean} on Line~5 is
a recursive definition, and so is \python{g.tfm} on Line~6,
implementing a search space with sub-pipelines of unbounded length.
While AlphaD3M uses reinforcement learning to search over this
grammar, Figure~\ref{fig:example_alphad3m} does something far less
sophisticated. Line~13 unfolds the grammar to depth~3, obtaining a
bounded planned pipeline, and Line~14 searches that using hyperopt,
with no further modifications required.

\begin{figure}
\begin{lstlisting}[language=python,frame=single,framerule=0pt,backgroundcolor=\color{background}]
g = Grammar()

g.start    = g.process >> g.features >> g.model
g.process  = g.process1 | ((g.process1 & g.process) >> Concat)
g.features = g.feature1 \
           | (((g.feature1 | g.est) & g.features) >> Concat)
g.model    = g.est

g.process1 = NoOp | MinMax | Standard | Norm | Robust
g.feature1 = NoOp | PCA | PolynomialFeatures | Nystroem
g.est      = GaussianNB | GradientBoostingClassifier | KNN \
           | RandomForestClassifier | ExtraTreesClassifier \
           | QDA | PassiveAggressiveClassifier \
           | DecisionTreeClassifier | LR | XGB | LGBM | SVC
\end{lstlisting}
\caption{\label{fig:example_tpot}Example grammar inspired by TPOT.}
\end{figure}

Figure~\ref{fig:example_tpot} shows a \lale grammar inspired by
TPOT~\cite{olson_et_al_2016}. It describes possibly \emph{non-linear}
pipelines, which use not only \python{>>} but also the~\python{&}
combinator. Recall that \python{(x & y) >> Concat} applies both
\python{x} and \python{y} to the same data and then concatenates the
features from both. Besides transformers, Line~6 also uses estimators
with~\python{&}, turning their predictions into features for
downstream operators.  This is not supported in scikit-learn pipelines
but is supported in \lale.

%% \begin{figure}
%% \lstinputlisting[language=python,frame=single,framerule=0pt,backgroundcolor=\color{background}]{example_ho_grammar.py}
%% \caption{\label{fig:example_ho_grammar}Example grammar with higher-order operator.}
%% \end{figure}

%% Figure~\ref{fig:example_ho_grammar} demonstrates how higher-order
%% operators and grammars work together. Line~6 uses the higher-order
%% \python{BaggingClassifier} operator, passing the non-terminal
%% \python{g.unstable} as a nested operator.

\emph{Progressive disclosure} is a design technique that makes things
easier to use by only requiring users to learn new features when and
as they need them. The starting point for \lale is manual machine
learning, and thus, the scikit-learn code in
Figure~\ref{fig:example_manual} is also valid \lale code. User needs
to learn zero new features if they do not use AutoML.  To use
algorithm selection, users only need to learn about the \python{|}
combinator and the \python{auto_configure} function.  To express
pipelines more concisely, users can learn about the \python{>>} and
\python{&} combinators, but those are optional syntactic sugar for
\python{make_pipeline} and \python{make_union} from scikit-learn.  To
use hyperparameter tuning, users only need to learn about
\python{wrap_imported_operators}. To exercise more control over the
search space, users can learn about freeze and custom schemas.  While
schemas are a non-trivial concept, \lale expresses them in JSON
Schema~\cite{pezoa_et_al_2016}, which is a widely-adopted and
well-documented standard proposal. To use higher-order operators,
users need not learn new syntax, as \lale supports scikit-learn syntax
for them. Finally, to use grammars, users need to add `\python{g.}' in
front of their pipeline definitions; however, all the other features,
such as the \python{|} combinator and the \python{auto_configure}
function, continue to work the same with or without grammars.

%%% Local Variables:
%%% mode: latex
%%% TeX-master: "main"
%%% End:

\section{Search Space Generation}\label{sec:compiler}

This section describes how to map the programming model from
Section~\ref{sec:language} to work directly with three popular AutoML
tools: scikit-learn's GridSearchCV~\cite{buitinck_et_al_2013},
hyperopt~\cite{bergstra_et_al_2015}, and
SMAC~\cite{hutter_hoos_leytonbrown_2011}, the library behind
auto-sklearn~\cite{feurer_et_al_2015}.

\subsection{From Grammars to Planned Pipelines}\label{sec:comp_to_planned}

\lale offers two approaches for using a grammar with Grid\-Search\-CV,
hyperopt, and SMAC: unfolding and sampling.  Both approaches produce a
planned pipeline, which can be directly used as the input for the
compiler in Section~\ref{sec:comp_from_planned}. Unfolding and
sampling are merely intended as proof-of-concept baseline
implementations. In the future, we will also explore integrating \lale
grammars directly with AutoML tools that support them, such as
AlphaD3M~\cite{drori_et_al_2019}.

Unfolding first expands the grammar to a given depth, such as~3 in the
example from Figure~\ref{fig:example_alphad3m} Line~13. Then, it
prunes all disjuncts containing unresolved nonterminals, so that only
planned \lale operators (individual, pipeline, or choice) remain.

Sampling traverses the grammar by following each nonterminal, picking
a random step in each choice, and unfolding each pipeline. The result
is a planned pipeline without any operator choices.

\subsection{From Planned Pipelines to Existing Tools}\label{sec:comp_from_planned}

This section sketches how to map a planned pipeline (which includes a
topology, steps for operator choices, and schemas for individual
operator) to a search space in the format required by
Grid\-Search\-CV, hyperopt, or SMAC. The running example for this
section is the pipeline
\python{PCA >> (J48 | LR)} with the individual operator schemas in
Figure~\ref{fig:example_hpschemas}.

\lale's search space generator has two phases: normalizer and
backend. The normalizer translates the schemas of individual operators
separately. The backend combines the schemas for the entire pipeline
and finally generates a tool-specific search space.

The \emph{normalizer} processes the schema for an individual operator
in a bottom-up pass. The desired end result is a search space in
\lale's \emph{normal form}, which is
\mbox{$\vee(\textrm{dict}\texttt{\{}\textrm{cat}^*,\textrm{cont}^*\texttt{\}}^*)$}.
At each level, the normalizer simplifies children and hoists
disjunctions up.
\[\begin{array}{@{}l@{}l@{\;}l}
  \textit{PCA} &:&\!\textrm{dict}\texttt{\{}N{:\,}(0..1)\texttt{\}} \vee \textrm{dict}\texttt{\{}N{:\,}[\textit{mle}]\texttt{\}}
\\
  \textit{J48}
&:&\!\textrm{dict}\texttt{\{}R{:\,}[\textit{false}],               C{:\,}(0..0.5)\texttt{\}} \vee
      \textrm{dict}\texttt{\{}R{:\,}[\textit{true},\textit{false}], C{:\,}[0.25]\texttt{\}}
\\
  \textit{LR}
&:&\!\textrm{dict}\texttt{\{}S{:\,}[\textit{linear}], P{:\,}[\textit{l1}, \textit{l2}]\texttt{\}} \vee
      \textrm{dict}\texttt{\{}S{:\,}[\textit{linear}, \textit{sag}, \textit{lbfgs}], P{:\,}[\textit{l2}]\texttt{\}}
\end{array}\]

The backend starts by first \emph{combining} the search spaces for all
operators in the pipeline. Each pipeline becomes a `dict' over its
steps; each operator choice becomes an `$\vee$' over its steps with
added discriminants $D$ to track what was chosen; and each individual
operator simply comes from the normalizer. This yields an intermediate
representation (\emph{IR}) whose nesting structure reflects the
operator nesting of the original pipeline. For our running example,
this is:
\[\textrm{dict}\left\{\begin{array}{@{}c@{\,}l@{}}
    0{:} & \textrm{dict}\texttt{\{}N{:\,}(0..1)\texttt{\}} \vee \textrm{dict}\texttt{\{}N{:\,}[\textit{mle}]\texttt{\}}\\
    1{:} & \left(\begin{array}{@{}l@{}}
             \,\left(\begin{array}{@{}l@{}}
               \textrm{dict}\texttt{\{}D{:\,}[\textit{J48}], R{:\,}[\textit{false}], C{:\,}(0..0.5)\texttt{\}} \;\vee\\
               \textrm{dict}\texttt{\{}D{:\,}[\textit{J48}], R{:\,}[\textit{true},\textit{false}], C{:\,}[0.25]\texttt{\}}
             \end{array}\right) \;\vee\\
             \left(\begin{array}{@{}l@{}}
               \textrm{dict}\texttt{\{}D{:\,}[\textit{LR}], S{:\,}[\textit{linear}], P{:\,}[\textit{l1}, \textit{l2}]\texttt{\}} \;\vee\\
               \textrm{dict}\texttt{\{}D{:\,}[\textit{LR}], S{:\,}[\textit{linear}, \textit{sag}, \textit{lbfgs}], P{:\,}[\textit{l2}]\texttt{\}}
             \end{array}\right)
           \end{array}\right)
\end{array}\right\}\]

The remainder of the backend is specific to the targeted AutoML tools,
which the following text describes one by one.

The \emph{hyperopt backend} of \lale is the simplest because hyperopt
supports nested search space specifications that are conceptually
similar to the \lale IR. For instance, an exclusive disjunction
`$\vee$' from the IR can be translated into a hyperopt
\python{hp.choice}, an example for which occurs in
Figure~\ref{fig:example_hyperopt} Line~5. Similarly, a `dict' from the
IR can be translated into a Python dictionary that hyperopt
understands. For working with higher-order operators, \lale adds
additional markers that enable it to reconstruct nested operators
later.

The \emph{SMAC backend} has to flatten \lale's nested IR into a grid
of disjuncts with discriminants~$D$.  To do this, it internally uses a
name mangling encoding that extends the \python{__} mangling of
scikit-learn, an example for which occurs in
Figure~\ref{fig:example_gridsearch} Line~7. Each element of the grid
needs to be a simple `dict' with no further nesting. For our
running example, the result in mathematical notation is:
\[\hspace*{-2mm}\begin{array}{c@{\,}l@{\,}l@{\,}l@{\,}l@{}l}
     & \textrm{dict}\texttt{\{}N{:\,}(0..1),         & D{:\,}[\textit{J48}], & R{:\,}[\textit{false}],                                & C{:\,}(0..0.5)                   & \texttt{\}}\\
\vee & \textrm{dict}\texttt{\{}N{:\,}(0..1),         & D{:\,}[\textit{J48}], & R{:\,}[\textit{true},\textit{false}],                  & C{:\,}[0.25]                     & \texttt{\}}\\
\vee & \textrm{dict}\texttt{\{}N{:\,}[\textit{mle}], & D{:\,}[\textit{J48}], & R{:\,}[\textit{false}],                                & C{:\,}(0..0.5)                   & \texttt{\}}\\
\vee & \textrm{dict}\texttt{\{}N{:\,}[\textit{mle}], & D{:\,}[\textit{J48}], & R{:\,}[\textit{true},\textit{false}],                  & C{:\,}[0.25]                     & \texttt{\}}\\
\vee & \textrm{dict}\texttt{\{}N{:\,}(0..1),         & D{:\,}[\textit{LR}],  & S{:\,}[\textit{linear}],                               & P{:\,}[\textit{l1}, \textit{l2}] & \texttt{\}}\\
\vee & \textrm{dict}\texttt{\{}N{:\,}(0..1),         & D{:\,}[\textit{LR}],  & S{:\,}[\textit{linear}, \textit{sag}, \textit{lbfgs}], & P{:\,}[\textit{l2}]              & \texttt{\}}\\
\vee & \textrm{dict}\texttt{\{}N{:\,}[\textit{mle}], & D{:\,}[\textit{LR}],  & S{:\,}[\textit{linear}],                               & P{:\,}[\textit{l1}, \textit{l2}] & \texttt{\}}\\
\vee & \textrm{dict}\texttt{\{}N{:\,}[\textit{mle}], & D{:\,}[\textit{LR}],  & S{:\,}[\textit{linear}, \textit{sag}, \textit{lbfgs}], & P{:\,}[\textit{l2}]              & \texttt{\}}
\end{array}\]

\noindent
Next, the SMAC backend adds conditionals that tell the Bayesian
optimizer which variables are relevant for which disjunct, and finally
outputs the search space in SMAC's PCS format.

The \emph{GridSearchCV backend} starts from the same flattened grid
representation that is also used by the SMAC backend. Then, it
discretizes each continuous hyperparameter into a categorical by first
including the default and then sampling a user-configurable number of
additional values from its range and prior distribution (such as
uniform in Figure~\ref{fig:example_hpschemas} Line~7).  The generated
search space in mathematical notation is:
\[\hspace*{-2mm}\begin{array}{c@{\,}l}
     & \textrm{dict}\texttt{\{}N{:\,}[0.50,0.01],    D{:\,}[\textit{J48}], R{:\,}[\textit{false}],                                C{:\,}[0.25, 0.01] \texttt{\}}\\
\vee & \textrm{dict}\texttt{\{}N{:\,}[0.50,0.01],    D{:\,}[\textit{J48}], R{:\,}[\textit{true},\textit{false}],                  C{:\,}[0.25]\texttt{\}}\\
\vee & \textrm{dict}\texttt{\{}N{:\,}[\textit{mle}], D{:\,}[\textit{J48}], R{:\,}[\textit{false}],                                C{:\,}[0.25, 0.01]\texttt{\}}\\
\vee & \textrm{dict}\texttt{\{}N{:\,}[\textit{mle}], D{:\,}[\textit{J48}], R{:\,}[\textit{true},\textit{false}],                  C{:\,}[0.25]\texttt{\}}\\
\vee & \textrm{dict}\texttt{\{}N{:\,}[0.50,0.01],    D{:\,}[\textit{LR}],  S{:\,}[\textit{linear}],                               P{:\,}[\textit{l1}, \textit{l2}]\texttt{\}}\\
\vee & \textrm{dict}\texttt{\{}N{:\,}[0.50,0.01],    D{:\,}[\textit{LR}],  S{:\,}[\textit{linear}, \textit{sag}, \textit{lbfgs}], P{:\,}[\textit{l2}]\texttt{\}}\\
\vee & \textrm{dict}\texttt{\{}N{:\,}[\textit{mle}], D{:\,}[\textit{LR}],  S{:\,}[\textit{linear}],                               P{:\,}[\textit{l1}, \textit{l2}]\texttt{\}}\\
\vee & \textrm{dict}\texttt{\{}N{:\,}[\textit{mle}], D{:\,}[\textit{LR}],  S{:\,}[\textit{linear}, \textit{sag}, \textit{lbfgs}], P{:\,}[\textit{l2}] \texttt{\}}
\end{array}\]

\section{Implementation}\label{sec:impl}

This section highlights some of the trickier parts of the \lale
implementation, which is entirely in Python.

To implement \emph{lifecycle states}, \lale uses Python subclassing.
For example, the \python{Trainable} is a subclass of \python{Planned},
adding a \python{fit} method. Subclassing lets users treat an operator
as also still belonging to an earlier state, e.g., in a mixed-state
pipeline. The \lale implementation adds Python~3 type hints so users
can get additional help from tools such as MyPy, PyCharm, or VSCode.

To implement the \emph{combinators} \python{>>}, \python{&}, and
\python{|}, \lale uses Python's overloaded \python{__rshift__},
\python{__and__}, and \python{__or__} methods. Python only supports
overriding these as instance methods.  Therefore, unlike in
scikit-learn, \lale planned operators are object instances, not
classes. This required emulating the scikit-learn \python{__init__}
with \python{__call__}.

The implementation carefully \emph{avoids in-place mutation} of
operators by methods such as \python{auto_configure}, \python{fit},
\python{customize_schema}, or \python{unfold}. This prevents
unintended side effects and keeps the implementation consistent with
the mathematical function abstractions from
Section~\ref{sec:lang_declarative}. Unfortunately, in scikit-learn,
\python{fit} does in-place mutation, so for compatibility, \lale
supports that but with a deprecation warning.

The implementation lets users \emph{import operators directly from
  their source packages}. For example, see
Figure~\ref{fig:lale_running_example} \mbox{Lines 2--5}. However,
these operators need to then support the combinators and have attached
schemas. \lale supports that via \python{wrap_imported_operators()},
which reflects over the symbol table and replaces any known non-\lale
operators by an object that points to the non-\lale operator and
augments it with \lale operator functionality.

The implementation supports \emph{interoperability with PyTorch, Weka,
  and R} operators. This is demonstrated by \lale's operators from
PyTorch (BERT, ResNet50), Weka (J48), and R~(ARulesCBA). Supporting
them requires, first, a class that is scikit-learn compatible. While
this is easy for some cases (e.g., XGBoost), it is sometimes
non-trivial. For instance, for Weka, \lale uses \python{javabridge}.
Second, each operator needs a JSON schema for its hyperparameters.
This is eased by \lale's \python{customize_schema} API.

The implementation of \emph{grammars} had to overcome the core
difficulty that recursive nonterminals require being able to use a
name before it is defined. Python does not allow that for local
variables. Therefore, \lale grammars implement it with object
attributes instead. More specifically, \lale grammars use overloaded
\python{__getattr__} and \python{__setattr__} methods.

\section{Results}\label{sec:results}

This section evaluates \lale on OpenML classification tasks and on different data modalities.
It also experimentally demonstrates the importance of side constraints for the optimization process.
For each experiment, we specified a \lale search space and then
used \python{auto_configure} to run hyperopt on it.
The value proposition of \lale is to leverage existing AutoML tools effectively and consistently; in general, we do not expect \lale to outperform them.

\subsection{Benchmarks: OpenML Classification}\label{sec:results_openml}

To demonstrate the use of \lale, we designed four experiments that
specify different search spaces for OpenML classification tasks.

\begin{figure}
\begin{lstlisting}[language=python,frame=single,framerule=0pt,backgroundcolor=\color{background}]
prep = ( NoOp | MinMaxScaler | StandardScaler | Normalizer 
       | RobustScaler )
feat = ( NoOp | PCA | PolynomialFeatures | Nystroem )
clf = ( GaussianNB | GradientBoostingClassifier | SVC
      | KNeighborsClassifier | RandomForestClassifier
      | ExtraTreesClassifier | QuadraticDiscriminantAnalysis
      | PassiveAggressiveClassifier | DecisionTreeClassifier
      | LogisticRegression | XGBClassifier | LGBMClassifier )
lale_openml_pipeline = prep >> feat >> clf
\end{lstlisting}
\caption{\label{fig:example_openml}Pipeline for OpenML experiment.}
\end{figure}

\begin{table*}\small
\caption{Accuracy for 15 OpenML classification tasks}
\label{tab:accuracy_openml}
\centerline{\begin{tabular}{@{}l|r@{ }l r@{ }l r@{ }l r@{ }l r@{ }l r@{ }l|rrrr@{}}
 & \multicolumn{12}{|c}{Absolute accuracy (mean and standard deviation over 5 runs)} & \multicolumn{4}{|c}{$100 * (\textit{accuracy}/\textsc{autoskl} - 1)$}\\
\textsc{Dataset} & \multicolumn{2}{c}{\textsc{autoskl}} &        \multicolumn{2}{c}{\textsc{lale-pipe}} & \multicolumn{2}{c}{\textsc{lale-tpot}} & \multicolumn{2}{c}{\textsc{lale-ad3m}} & \multicolumn{2}{c}{\textsc{lale-adb}} & \multicolumn{2}{c|}{\textsc{askl-adb}} & \textsc{lale-pipe} & \textsc{lale-tpot} & \textsc{lale-ad3m} & \textsc{lale-adb} \\
\midrule
australian    & 85.09 & (0.44) & 85.44 & (0.72) & 85.88 & (0.57) & 86.84 & (0.00) &  86.05 & (1.62) & 84.74 & (3.11)  &  0.41 &  0.93  &   2.06 &  1.13\\
blood         & 77.89 & (1.39) & 76.28 & (5.22) & 77.49 & (2.46) & 74.74 & (0.74) &  77.09 & (0.74) & 74.74 & (0.84)  & -2.08 & -0.52  &  -4.05 & -1.04\\
breast-cancer & 73.05 & (0.58) & 71.16 & (1.20) & 71.37 & (1.15) & 69.47 & (3.33) &  70.95 & (2.05) & 72.42 & (0.47)  & -2.59 & -2.31  &  -4.90 & -2.88\\
car           & 99.37 & (0.10) & 98.25 & (1.16) & 99.12 & (0.12) & 92.71 & (0.63) &  98.28 & (0.26) & 98.25 & (0.25)  & -1.13 & -0.25  &  -6.70 & -1.09\\
credit-g      & 76.61 & (1.20) & 74.85 & (0.52) & 74.12 & (0.55) & 74.79 & (0.40) &  76.06 & (1.27) & 76.24 & (1.02)  & -2.29 & -3.24  &  -2.37 & -0.71\\
diabetes      & 77.01 & (1.32) & 77.48 & (1.51) & 76.38 & (1.11) & 77.87 & (0.18) &  75.98 & (0.48) & 75.04 & (1.03)  &  0.61 & -0.82  &   1.12 & -1.33\\
hill-valley   & 99.45 & (0.97) & 99.25 & (1.15) & 100.0 & (0.00) & 96.80 & (0.21) & 100.00 & (0.00) & 99.10 & (0.52)  & -0.20 &  0.55  &  -2.66 &  0.55\\
jungle-chess  & 88.06 & (0.24) & 90.29 & (0.00) & 88.90 & (2.05) & 74.14 & (2.02) &  89.41 & (2.29) & 86.87 & (0.20)  &  2.54 &  0.96  & -15.80 &  1.53\\
kc1           & 83.79 & (0.31) & 83.48 & (0.75) & 83.48 & (0.54) & 83.62 & (0.23) &  83.30 & (0.36) & 84.02 & (0.31)  & -0.38 & -0.38  &  -0.21 & -0.58\\
kr-vs-kp      & 99.70 & (0.04) & 99.34 & (0.07) & 99.43 & (0.00) & 96.83 & (0.14) &  99.51 & (0.10) & 99.47 & (0.16)  & -0.36 & -0.27  &  -2.87 & -0.19\\
mfeat-factors & 98.70 & (0.08) & 97.58 & (0.28) & 97.18 & (0.50) & 97.55 & (0.07) &  97.52 & (0.40) & 97.94 & (0.08)  & -1.14 & -1.54  &  -1.17 & -1.20\\
phoneme       & 90.31 & (0.39) & 89.06 & (0.67) & 89.56 & (0.36) & 76.57 & (0.00) &  90.11 & (0.45) & 91.36 & (0.21)  & -1.39 & -0.83  & -15.20 & -0.22\\
shuttle       & 87.27 & (11.6) & 99.94 & (0.01) & 99.93 & (0.04) & 99.89 & (0.00) &  99.98 & (0.00) & 99.97 & (0.01)  & 14.51 & 14.50  &  14.45 & 14.56\\
spectf        & 87.93 & (0.86) & 87.24 & (1.12) & 88.45 & (2.25) & 83.62 & (6.92) &  88.45 & (2.63) & 89.66 & (2.92)  & -0.78 &  0.59  &  -4.90 &  0.59\\
sylvine       & 95.42 & (0.21) & 95.00 & (0.61) & 94.41 & (0.75) & 91.31 & (0.12) &  95.15 & (0.20) & 95.07 & (0.14)  & -0.45 & -1.07  &  -4.31 & -0.29\\
\end{tabular}}
\end{table*}

\begin{description}[style=unboxed,leftmargin=0em, itemsep=0.5em]
\item[lale-pipe:] The three-step planned \python{lale_openml_pipeline}
in Figure~\ref{fig:example_openml}.
\item[lale-ad3m:] The AlphaD3M-inspired grammar of Figure~\ref{fig:example_alphad3m} unfolded with a maximal depth of~3.
\item[lale-tpot:] The TPOT-inspired grammar of Figure~\ref{fig:example_tpot} unfolded with a maximal depth of~3.
\item[lale-adb:] The higher-order operator pipeline of Figure~\ref{fig:example_adaboost}.
\end{description}

For comparison, we used auto-sklearn~\cite{feurer_et_al_2015} --- a popular scikit-learn based AutoML tool that has won two OpenML challenges --- with its default setting as a baseline.
We chose 15 OpenML datasets for which we could get meaningful results (more than 30 trials) using the default settings of Auto-sklearn.
The selected datasets comprise 5 simple classification tasks (test accuracy~$>90\%$ in all our experiments) and 10 harder tasks (test accuracy~$< 90\%$).
For each experiment, we used a $66\%-33\%$ validation-test split, and a 5-fold cross validation on the validation split during optimization.
Experiments were run on a 32 cores (2.0GHz) virtual machine with 128GB memory, and for each task, the total optimization time was set to 1~hour with a timeout of 6~minutes per trial.

% As a baseline for comparison with lale-adb, we used Auto-sklearn with only data preprocessing and restricting the classifier to AdaBoostClassifier.
% The preprocessing options in lale-adb were chosen to be same as the Auto-sklearn data preprocessing for a fair comparison.

Table~\ref{tab:accuracy_openml} presents the results of our experiments.
For each experiment, we report the test accuracy of the best pipeline
found averaged over 5 runs.
Note that for the \emph{shuttle} dataset, 3 out of 5 runs of auto-sklearn resulted in a \python{MyDummyClassifier} being returned as the result. Since we were
trying to evaluate the default settings, we did not attempt debugging it, but according to the tool's issue log, other users have encountered it before.
The column \emph{askl-adb} reports the accuracy of auto-sklearn when the set of classifiers is limited to AdaBoost with only data pre-processing.
These results are presented for comparison with the \emph{lale-adb} experiments, as the data preprocessing operators in Figure~\ref{fig:example_adaboost} 
were chosen to match those of auto-sklearn as much as possible.
Also note that the default setting for auto-sklearn uses meta-learning.

The results show that carefully crafted search spaces (e.g., TPOT-inspired grammar, or pipelines with higher-order operators) and off-the-shelf optimizers such as hyperopt can achieve accuracies that are competitive with state-of-the-art tools.
These experiments thus validate \lale's controlled approach to AutoML as an alternative to black-box solutions.
In addition, these experiments illustrate that \lale is modular enough to easily express and compare multiple search spaces for a given task.

\subsection{Case Studies: Other Modalities}\label{sec:results_modalities}

While all the examples so far focused on tasks for tabular datasets, the core contribution of \lale is not limited to those. 
This section demonstrates \lale's versatility on three datasets from different modalities.
Table~\ref{tab:modalities_results} summarizes the results.

\begin{table}
  \caption{\label{tab:modalities_results}Performance of the best
    pipeline using \lale with hyperopt. The mean and stdev are over 3 runs.}
  \centerline{\small\begin{tabular}[t]{@{}llrlrl@{}}
    \textsc{modality}    & \textsc{dataset}                   & \textsc{mean} & \textsc{stdev} &  \textsc{metric}\\
    \midrule
    Text        & Drug Review   & 1.9237 & 0.06 & test RMSE\\
    Image       & CIFAR-10   & 93.53\% & 0.11 & test accuracy \\
    Time-series & Epilepsy  & 73.15\% & 8.2 & test accuracy\\
  \end{tabular}}
\end{table}

\begin{description}[style=unboxed,leftmargin=0em, itemsep=0.5em]
  
\item[Text.]
We used the Drug Review dataset for predicting a rating  given by a patient to a drug. 
The Drug Review dataset has a text column called \python{review}, to which the pipeline applies either \python{TfidfVectorizer} from scikit-learn or a pretrained 
\python{BERT} embeddings, which is a text embedding based on neural networks. 
The dataset also has numeric columns, which are concatenated with the result of the embedding.

\begin{lstlisting}[language=python,frame=single,framerule=0pt,backgroundcolor=\color{background}]
planned_pipeline = (
  Project(columns=['review']) >> (BERT | TfidfVectorizer)
  & Project(columns={'type': 'number'})
) >> Cat >> (LinearRegression | XGBRegressor)  
\end{lstlisting}

\item[Image.]
We used the CIFAR-10 computer vision dataset.
We picked the \python{ResNet50} deep-learning model, since it has been shown to do well on CIFAR-10.  
Our experiments kept the architecture of \python{ResNet50} fixed and tuned learning-procedure hyperparameters.

\item[Time-series.]
We used the Epilepsy dataset, a subset of the TUH Seizure Corpus, for
classifying seizures by onset location (generalized or focal).  We
used a three-step pipeline:

\begin{lstlisting}[language=python,frame=single,framerule=0pt,backgroundcolor=\color{background}]
planned = Window \
  >> (KNeighborsClassifier| XGBClassifier| LogisticRegression) \
  >> Voting
\end{lstlisting}

We implemented a popular pre-processing method~\cite{Schindler2006} in
a \python{Window} operator with three hyperparameters $W$,
$O$, and~$T$. Note that this transformer leads to multiple samples per seizure.
Hence, during evaluation, each seizure is classified by taking a vote
of the predictions made by each sample generated from it.
\end{description}

\begin{figure}
\includegraphics[width=\columnwidth]{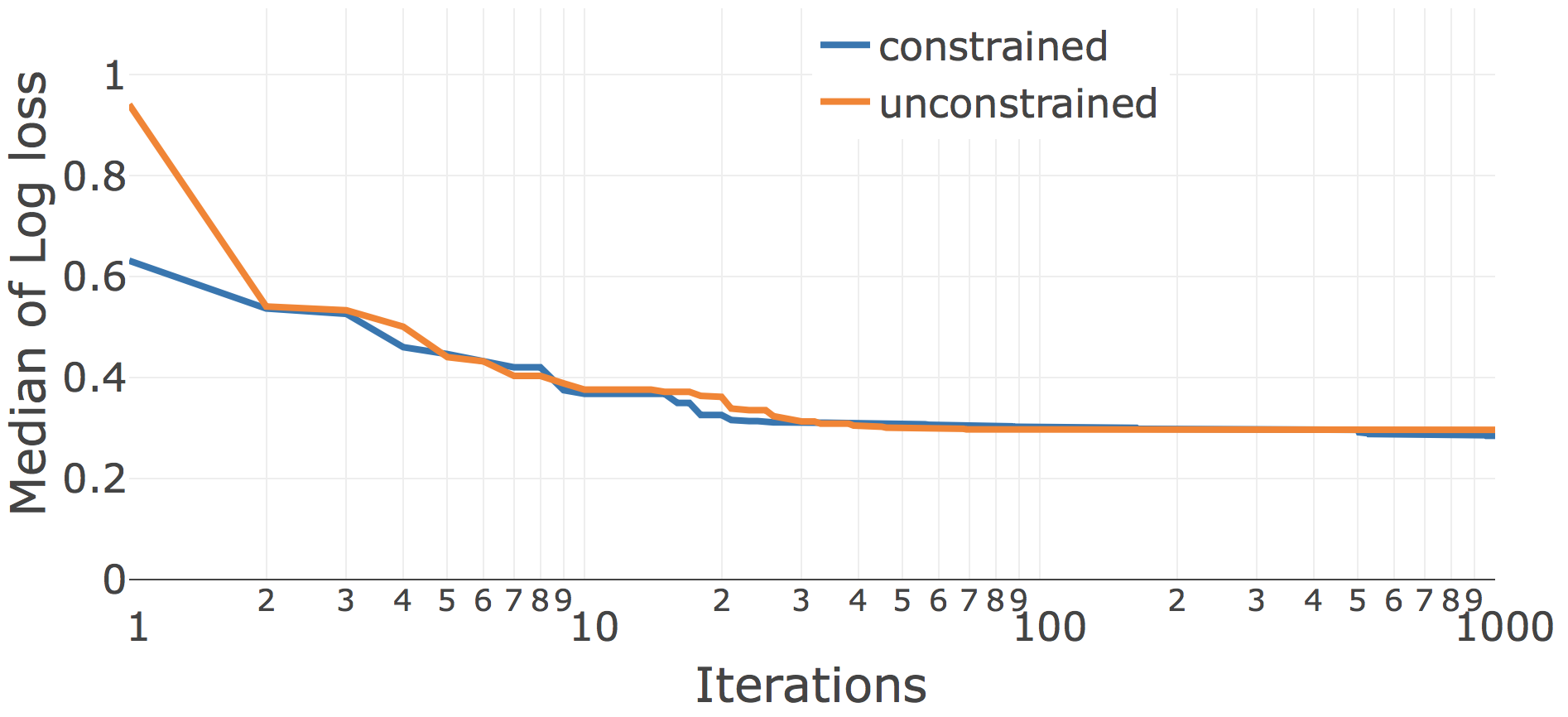}
\caption{\label{fig:car_2classifiers}Convergence with planned pipeline \python{LR | KNN}.}
\vspace*{3mm} 
\includegraphics[width=\columnwidth]{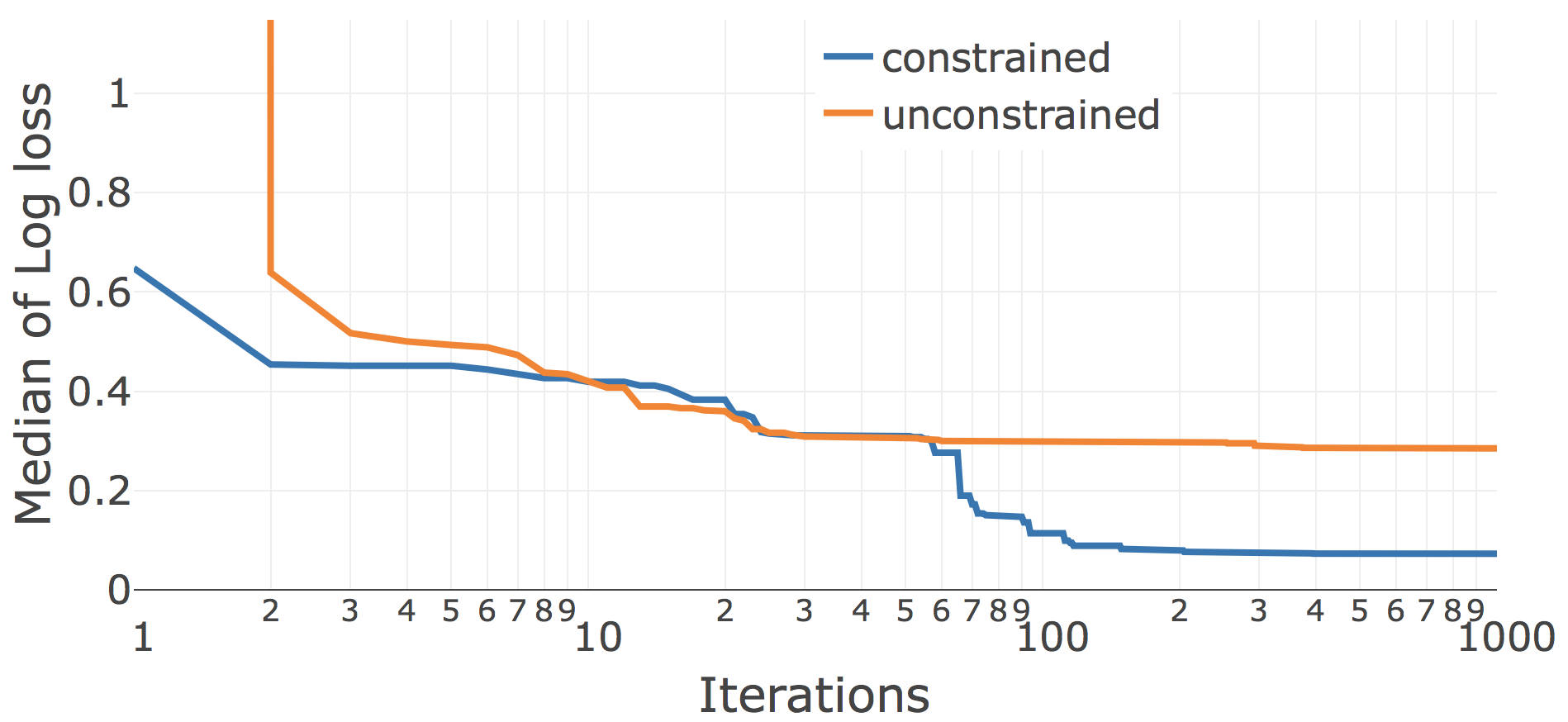}
\caption{\label{fig:car_3classifiers}Convergence with planned pipeline \python{J48 | LR | KNN}.}
\end{figure}

\subsection{Effect of Side Constraints on Convergence}\label{sec:results_convergence}
\lale's search space compiler takes rich hyperparameter schemas
including side constraints and translates them into semantically
equivalent search spaces for different AutoML tools. This raises the
question of how important those side constraints are in practice.  To
explore this, we did an ablation study where we generated not just the
\emph{constrained} search spaces that are default with \lale but also
\emph{unconstrained} search spaces that drop side constraints. With
hyperopt on the unconstrained search space, some iterations are
unsuccessful due to exceptions, for which we reported
\python{np.float.max} loss.
Figure~\ref{fig:car_2classifiers} plots the convergence for the Car
dataset on the planned pipeline \python{LR | KNN}. Both of these
operators have a few side constraints. Whereas the unconstrained
search space causes some invalid points early in the search, the two
curves more-or-less coincide after about two dozen iterations.
% elided ARules, because it doesn't have predict_proba needed for log-loss
The story looks very different in Figure~\ref{fig:car_3classifiers}
when adding a third operator \python{J48 | LR | KNN}.  In the
unconstrained case, \python{J48} has many more invalid runs, causing
hyperopt to see so many \python{np.float.max} loss values from
\python{J48} that it gives up on it. In the constrained case, on the
other hand, \python{J48} has no invalid runs, and hyperopt eventually
realizes that it can configure \python{J48} to obtain substantially
better performance.

\subsection{Dataset Details}
In order to be specific about the exact datasets for reproducibility,
Table~\ref{tab:datasets} report the URLs for accessing those.

\begin{table}\small
\caption{\label{tab:datasets}Dataset details for reproducibility.}
\begin{center}
\begin{tabular}{@{}lp{23em}@{}}
\textsc{Dataset} & \textsc{URL} \\\midrule
australian	    &  \url{https://www.openml.org/d/40981} \\
blood	        &  \url{https://www.openml.org/d/1464} \\
breast-cancer	&  \url{https://www.openml.org/d/13}\\
car	            &  \url{https://www.openml.org/d/40975} \\
credit-g	    &  \url{https://www.openml.org/d/31} \\
diabetes	    &  \url{https://www.openml.org/d/37} \\
hill-valley	    &  \url{https://www.openml.org/d/1479}\\
jungle-chess	&  \url{https://www.openml.org/d/41027} \\
kc1	            &  \url{https://www.openml.org/d/1067} \\
kr-vs-kp	    &  \url{https://www.openml.org/d/3} \\
mfeat-factors	&  \url{https://www.openml.org/d/12} \\
phoneme	        &  \url{https://www.openml.org/d/1489} \\
shuttle	        &  \url{https://www.openml.org/d/40685} \\
spectf	        &  \url{https://www.openml.org/d/337}\\
sylvine	        &  \url{https://www.openml.org/d/41146}\\\midrule
CIFAR-10        &  \url{https://en.wikipedia.org/wiki/CIFAR-10} \\
Drug Review     & \url{https://archive.ics.uci.edu/ml/datasets/Drug+Revi+Dataset+%28Drugs.com%29} \\
Epilepsy        & \url{https://www.ncbi.nlm.nih.gov/pmc/articles/PMC6246677} \\
\end{tabular}
\end{center}
\end{table}

%% \begin{alltt}\textcolor{red}{TODO}\scriptsize
%% - hyperopt ran for 1000 evaluations and found J48 with:
%%     dict\{'A': True, 'B': False, 'C': 0.25, 'L': False, 'M': 1,
%%           'N': 3, 'R': False, 'S': False, 'U': True\}
%% - GridSearchCV ran for 960 (there is no way to control the exact number 
%%   of grid points in our grid search implementation) and found J48 with:
%%     dict\{'A': False, 'B': False, 'C': 0.25, 'L': False, 'M': 1,
%%           'N': 3, 'R': False, 'S': False, 'U': False\}
%% - both of the above classifiers lead to the same test accuracy (98.07\%)
%%   and similar 10-fold cv accuracies: 86.54\% and 85.96\% respectively.
%% - SMAC \textcolor{red}{TODO: Get SMAC to run for 1000 iterations.}
%% \end{alltt}

\section{Related Work}\label{sec:related}

%% \begin{alltt}\textcolor{red}{TODO}\scriptsize
%% - page budget: 0.5 pages for this section + 0.4 pages for references
%% - search tools
%%   - scikit-learn GridSearchCV \cite{buitinck_et_al_2013}
%%   - Hyperopt \cite{bergstra_et_al_2015}
%%   - hyperopt-sklearn \cite{komer_bergstra_eliasmith_2014}
%%   - SMAC \cite{hutter_hoos_leytonbrown_2011}
%%   - auto-sklearn \cite{feurer_et_al_2015}
%%   - auto-Weka \cite{thornton_et_al_2013}
%% - schemas and types
%%   - Weka \cite{hall_et_al_2009}
%%   - pandas \cite{mckinney_2011}
%%   - data schemas \cite{breck_et_al_2019}
%%   - strongly typed genetic programming \cite{pilat_kren_neruda_2016}
%%   - ML Bazaar \cite{smith_et_al_2019}
%% - grammars
%%   - Recipe \cite{desa_et_al_2017}
%%   - AlphaD3M \cite{drori_et_al_2019}
%%   - TPOT \cite{olson_et_al_2016}
%% \end{alltt}

There are various search tools, designed around
scikit-learn~\cite{buitinck_et_al_2013}, and each usually focused on a
particular novel optimization algorithm. Auto-sklearn~\cite{feurer_et_al_2015}
uses the search space specification and optimization algorithm of
SMAC~\cite{hutter_hoos_leytonbrown_2011}.
Hyperopt-sklearn~\cite{komer_bergstra_eliasmith_2014} uses its own search space
specification scheme and a novel optimizer based on Tree-structured Parzen
Estimators (TPE)~\cite{bergstra_et_al_2011}. Scikit-learn also comes with its
own GridSearchCV and RandomizedSearchCV classes.
Auto-Weka~\cite{thornton_et_al_2013} is a predecessor of auto-sklearn that also
uses SMAC but operates on operators from Weka~\cite{hall_et_al_2009} instead of
scikit-learn.  TPOT~\cite{olson_et_al_2016} is designed around scikit-learn and
uses genetic programming to search for pipeline topologies and operator
choices. This usually leads to the generation of many misconfigured pipelines,
wasting execution time. RECIPE~\cite{desa_et_al_2017} prunes away the
misconfigurations to save execution time by validating generated pipelines with
a grammar for linear pipelines. AlphaD3M~\cite{drori_et_al_2019} makes use of a
grammar in a generative manner (instead of just for validation) with a deep
reinforcement learning based algorithm. Katz et al.~\cite{katz_et_al_2020}
similarly use a grammar for \lale pipelines with AI planning tools. In
contrast to these tools, the contribution of this paper is not a novel
optimization algorithm but rather a more consistent programming model with
search space generation targeting existing tools.

%% \begin{alltt}\textcolor{red}{TODO}\scriptsize
%% - Not sure what to discuss regarding pandas, data schemas, typed GP and ML Bazaar
%% \end{alltt}

\section{Conclusion}\label{sec:concl}

This paper describes \lale, a library for semi-automated data science.
\lale contributes a syntax that is consistent with scikit-learn, but
extends it to support a broad spectrum of automation including
algorithm selection, hyperparameter tuning, and topology search.
\lale works by automatically generating search spaces for established
AutoML tools, extending their capabilities to grammar-based search and
to search inside higher-order operators. The experiments show that search spaces crafted using 
\lale achieve results that are competitive with state-of-the-art tools while offering more versatility.

\newpage
\balance
\bibliography{bibfile}

\end{document}